\documentclass[10pt,twocolumn,letterpaper]{article}

\usepackage{iccv}              

\definecolor{iccvblue}{rgb}{0.21,0.49,0.74}
\usepackage[pagebackref,breaklinks,colorlinks,allcolors=iccvblue]{hyperref}

\def\paperID{Workshop Short non-archival paper} 
\def\confName{ICCV}
\def\confYear{2025}

\title{Morphological Analysis of Semiconductor Microstructures using Skeleton Graphs}

\author{Noriko Nitta\\
Kochi University of Technology\\
Kami, Kochi, Japan\\
{\tt\small nitta.noriko at kochi-tech.ac.jp}
\and
Rei Miyata\\
Kochi University of Technology\\
Kami, Kochi, Japan\\
\and
Naoto Oishi\\
National Institute of Technology, Kochi College \\
Nankoku, Kochi, Japan\\
{\tt\small oishi at kochi-ct.ac.jp}
}

\begin{document}
\maketitle
\begin{abstract}
In this paper, electron microscopy images of microstructures formed on Ge surfaces by ion beam irradiation were processed to extract topological features as skeleton graphs, which were then embedded using a graph convolutional network. The resulting embeddings were analyzed using principal component analysis, and cluster separability in the resulting PCA space was evaluated using the Davies-Bouldin index. The results indicate that variations in irradiation angle have a more significant impact on the morphological properties of Ge surfaces than variations in irradiation fluence.
\end{abstract}    
\section{Introduction}
\label{sec:intro}

Electron microscopy images are necessary for observing microstructures and are widely used in materials science and nanotechnology. Extracting shape features from these images remains a significant challenge for improving classification and pattern recognition accuracy. Traditionally, electron microscopy image classification has relied on geometric features (e.g., diameter, perimeter, and area) and texture features (e.g., intensity histograms, gray level co-occurrence matrix (GLCM) \cite{haralick1973textural}, and local binary pattern (LBP) \cite{ojala2002multiresolution}). More recently, feature extraction methods based on deep neural networks (e.g., ResNet \cite{he2016deep} and VGG \cite{Simonyan2015vgg}) and topological techniques (such as persistent homology \cite{edelsbrunner2002topological}) have been explored. Although these approaches capture global shape differences, they still struggle to represent detailed topological information.

In this study, topological features are extracted from electron microscopy images as skeleton graphs \cite{blum1967transformation}, which are then embedded using a graph convolutional network (GCN) \cite{kipf2017semi}, and the resulting topological descriptors are analyzed using principal component analysis (PCA) \cite{Pearson1901} to perform image classification. The electron microscopy images used here capture the microstructures formed on the surface of germanium (Ge), a semiconductor material, by ion beam irradiation. When Ge is irradiated with heavy ion beam, microstructures are formed on its surface \cite{oishi2018nanoporous}. Since microstructure morphologies exhibit diverse patterns depending on irradiation conditions, understanding the relationship between morphology and ion beam parameters will provide useful insights for process informatics.

Although skeleton graphs have proven effective for pose estimation \cite{cao2017openpose} and motion analysis \cite{yan2018spatial}, their use in extracting features from electron microscopy images remains largely unexplored. In research applying skeleton graphs to the structure of organic photovoltaics, features such as the number of skeletal pixels and endpoints on the skeleton have been extracted and used for machine learning predictions \cite{Jivani2022skeletal}. Similar graph neural network (GNN)-based methods have been used to analyse microstructural data in materials science, including nanoporous gold \cite{hu2016nanoporousgold}, crystal grains \cite{Thomas2023}, and nanotubes \cite{Vecchio2021}. This approach produces sparse embeddings that reduce computational costs and enhance classification accuracy by capturing detailed topological information using skeleton graphs. We focus on applying this technique to the analysis of electron microscopy images and the formation of microstructures induced by ion beam irradiation.
\section{Experimental Procedure}
\label{sec:formatting}

\begin{figure*}[htb]
  \centering
  \includegraphics[width=0.8\textwidth]{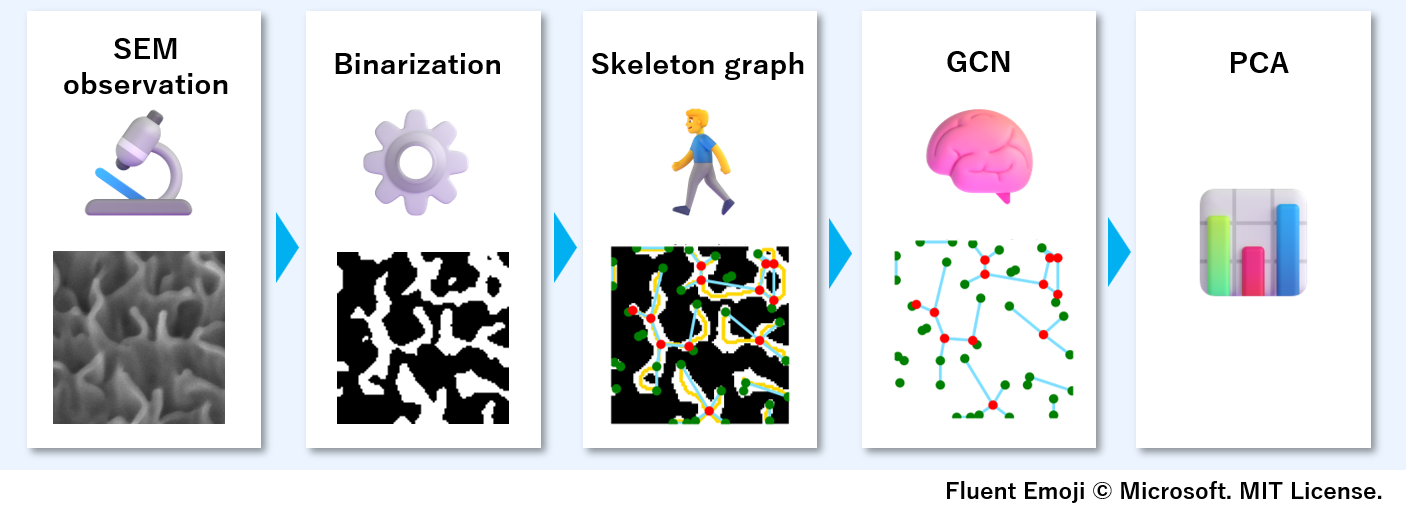}
  \caption{Schematic illustration of the experimental pipeline.}
  \label{fig:short}
\end{figure*}

Figure 1 illustrates the experimental pipeline used in this study. Single-crystal Ge was used as the sample material, and structures were fabricated by ion beam irradiation using a focused ion beam (FIB; FEI QUANTA3D200i). The irradiation parameters were as follows: ion species, Ga$^+$; acceleration voltage, 30 kV; fluences, $2\times10^{20}$, $5\times10^{20}$, and $1\times10^{21}\,\mathrm{ions}/\mathrm{m}^2$; and irradiation angles, $0^\circ$, $30^\circ$, and $45^\circ$. Structural evaluations were conducted via scanning electron microscopy (SEM; HITACHI SU8020) for surface observations, with an acceleration voltage of 5 kV and secondary electron detection. SEM images were preprocessed using OpenCV \cite{bradski2000opencv}. A Gaussian blur was applied to reduce noise, followed by binarization and inversion with a threshold of 100. Skeletonization of the binary images was performed using scikit-image \cite{vandewalt2014scikitimage}, \cite{zhang_suen_1984}. The resulting skeletons were converted into graph representations consisting of nodes and edges using the sknw library \cite{wang2018imagepy}, \cite{yxdragon_sknw_2025}.

The binary image \(B\) was first skeletonized by applying the morphological operator \(\mathcal{S}\), yielding the skeleton image (yellow)
\[
  S = \mathcal{S}(B).
\]
From \(S\), we constructed the undirected graph \(G=(V,E)\), where
\[
\begin{aligned}
  V &= \{\,p \mid S(p)=1 \land \deg(p)\ge1\},\\
  E &= \{(p,q)\mid p,q\in V,\;q\text{ is adjacent to }p\}.
\end{aligned}
\]
Each node \(n\) was annotated with its coordinate vector \(\mathbf{o}_n=(i_n,j_n)\) and classified by degree: nodes with \(\deg(n)\ge3\) were plotted in red to indicate branching points, while those with \(\deg(n)\le2\) were plotted in green to indicate endpoints. Finally, a figure was generated using Matplotlib \cite{hunter2007matplotlib}, displaying \(B\) in the binary image and overlaying the skeleton pixels \(p\) and \(q\) , the coded nodes at \(\mathbf{o}_n\), and straight line segments between \(\mathbf{o}_u\) and \(\mathbf{o}_v\) for each edge \((u,v)\in E\), thus clearly depicting the resulting skeleton graph structure. 

A GCN model was then constructed using PyTorch Geometric \cite{fey2019fast} to generate embeddings from the skeleton graphs. The GCN architecture comprised two GCNConv layers with hidden dimensions of 32, each followed by a ReLU activation, and a global mean pooling layer to produce a 32-dimensional graph-level embedding. Each node was initialized with a single constant scalar feature of 1. Subsequently, the extracted embeddings were subjected to PCA using scikit-learn \cite{pedregosa2011scikit}. The quality of PCA-based cluster separation was quantified using the Davies-Bouldin index (DBI) \cite{davies1979cluster}. For \(K\) clusters, the DBI is defined as:
\[
\mathrm{DBI}
= \frac{1}{K} \sum_{i=1}^{K} \max_{j \neq i}
\left(
\frac{S_i + S_j}{d(\mu_i, \mu_j)}
\right),
\]
where
\[
S_i = \frac{1}{\lvert C_i \rvert} \sum_{x \in C_i} \lVert x - \mu_i \rVert,
\quad
d(\mu_i, \mu_j) = \lVert \mu_i - \mu_j \rVert.
\]

Here,
\begin{itemize}
  \item $C_i$ and $C_j$ are the sets of points in clusters $i$ and $j$, respectively,
  \item $\mu_i$ and $\mu_j$ are the centroids of clusters $i$ and $j$, respectively,
  \item $S_i$ and $S_j$ denote the within-cluster scatters for clusters $i$ and $j$, and
  \item $d(\mu_i,\mu_j)$ is the between-cluster distance (the Euclidean distance between $\mu_i$ and $\mu_j$).
\end{itemize}

\begin{figure}[t]
  \centering
   \includegraphics[width=1\linewidth]{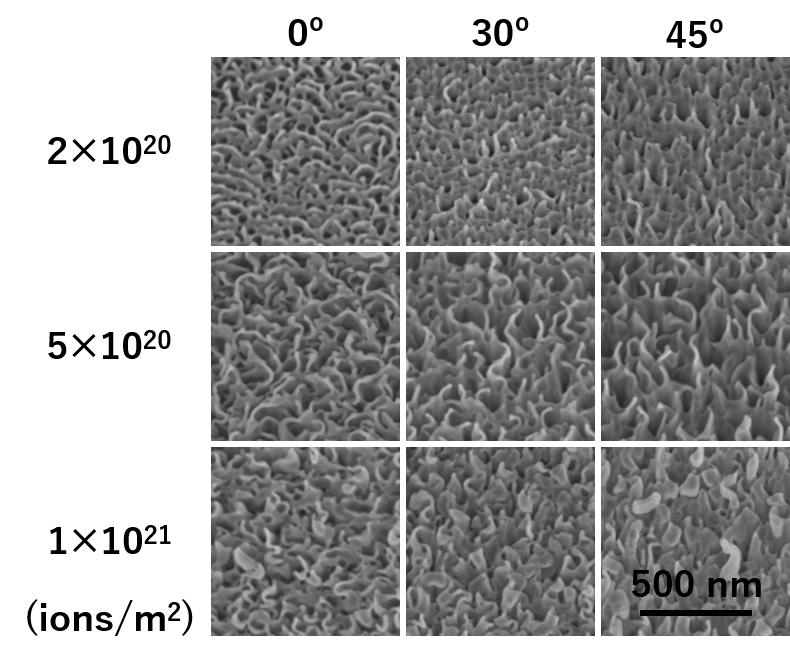}

   \caption{SEM images of ion beam-irradiated Ge surface structures.}
   \label{fig:onecol}
\end{figure}

\section{Results and Discussion}

The microstructures formed on the surfaces induced by ion beam irradiation of Ge are shown in Fig. 2 for fluences of $2\times10^{20}$, $5\times10^{20}$, and $1\times10^{21}\,\mathrm{ions}/\mathrm{m}^2$ at irradiation angles of $0^\circ$, $30^\circ$, and $45^\circ$. As the fluence increases, the fine features coarsen, and as the irradiation angle increases, the structures become progressively oblique. To quantify the topology of these structures, skeleton trajectories  (yellow) were first overlaid the binarized SEM images of the wall regions, from which branching points (red) and endpoints (green) were identified and connected to form a GNN (Fig. 3). The same pipeline was then applied to the inverted image to capture the morphology of the hole, resulting in a second GNN (Fig. 4).
The GCN-embedded features extracted from these skeleton graphs were then analyzed using PCA. In the PCA projection shown in Fig. 5, the points with the highest principal component scores for each class were connected to form a convex hull. The separation between the wall and hole classes was poor (DBI = 13.2), indicating significant overlap in their topological features. Variations in irradiation fluence and angle were assessed for their effects on the structural embeddings of wall and hole features, both jointly and separately. As shown in Fig. 6, fluence-dependent PCA yielded DBI values of 6.16 (walls + holes), 19.0 (walls only), and 3.14 (holes only), whereas angle-dependent PCA achieved significantly better separability, with DBI values of 2.35 (walls + holes), 1.23 (walls only), and 3.43 (holes only). These findings demonstrate that varying the irradiation angle captures morphological distinctions more effectively than changing the fluence. However, despite the clear visual differences in the images, the effect of fluence has not been sufficiently captured, which remains an issue for future investigation. In future work, we plan to compare the proposed approach with traditional texture-based, and neural network descriptors. We also aim to increase the sample size and, using the features embedded from skeleton graphs, apply supervised learning methods such as logistic regression and Gaussian process regression.

\begin{figure}[t]
  \centering
   \includegraphics[width=0.7\linewidth]{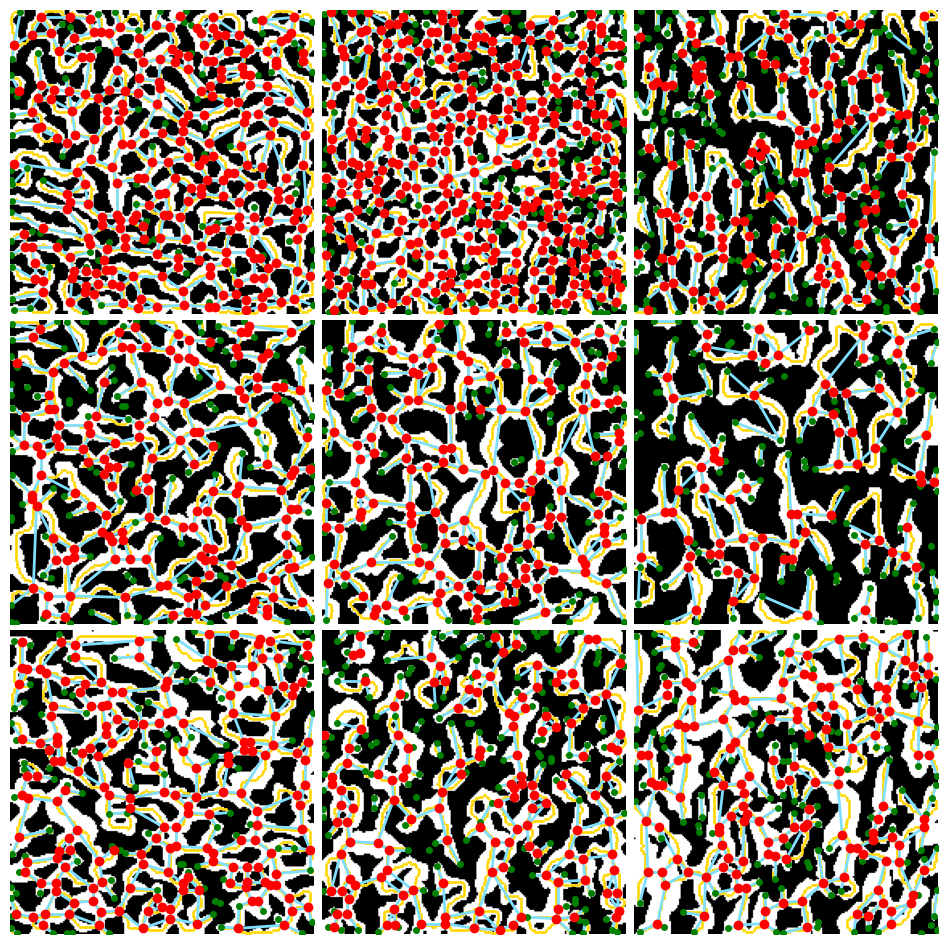}

   \caption{Skeleton graph construction for wall features following binarization of SEM images.}
   \label{fig:onecol1}
\end{figure}

\begin{figure}[t]
  \centering
   \includegraphics[width=0.7\linewidth]{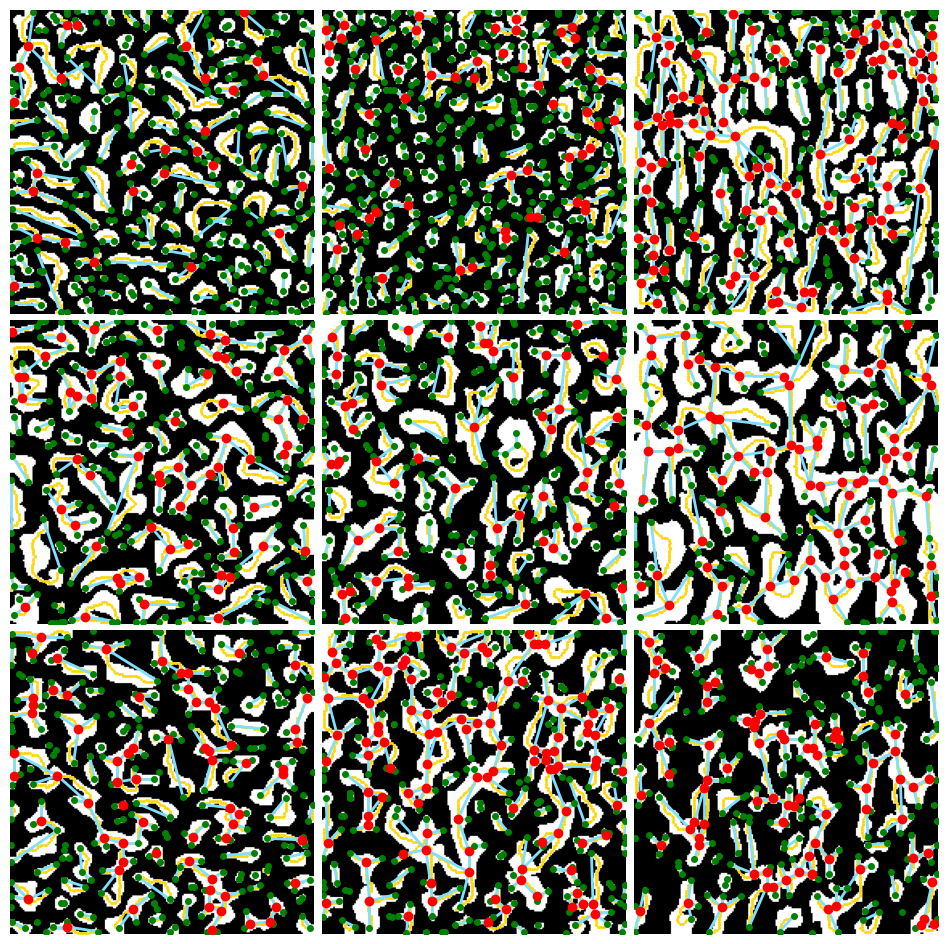}

   \caption{Skeleton graph construction for hole features after binarization of SEM images.}
   \label{fig:onecol2}
\end{figure}

\begin{figure}[t]
  \centering
   \includegraphics[width=0.9\linewidth]{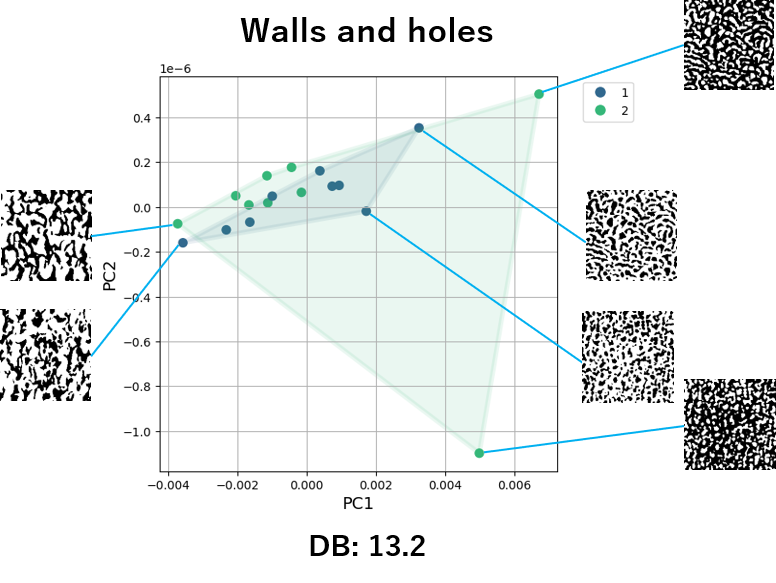}

   \caption{PCA of wall and hole structures (1: wall, 2: hole). The images in this figure are re-presented as binary images with the GNN removed from  those in Fig. 3 and Fig. 4.}
   \label{fig:onecol3}
\end{figure}

\begin{figure}[t]
  \centering
  \includegraphics[width=\linewidth]{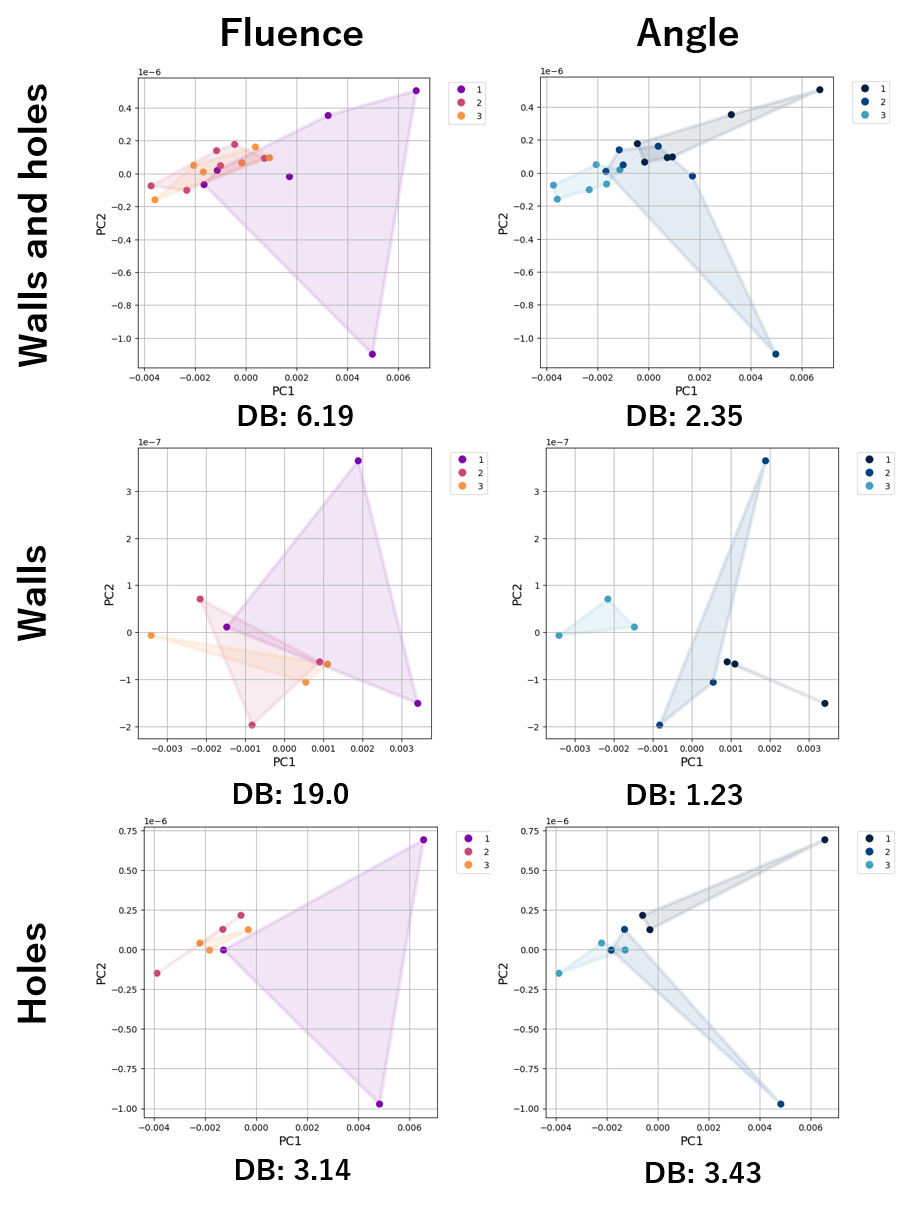}
  \caption{PCA of structural variations in walls and holes in relation to irradiation fluence (1: $2\times10^{20}\,\mathrm{ions}/\mathrm{m}^2$, 2: $5\times10^{20}\,\mathrm{ions}/\mathrm{m}^2$, 3: $1\times10^{21}\,\mathrm{ions}/\mathrm{m}^2$) and irradiation angle (1: $0^\circ$, 2: $30^\circ$, 3: $45^\circ$).The top two images in this figure are re-presentations of Fig. 5.}
  \label{fig:onecol4}
\end{figure}

\section{Conclusion}

The topological features of the Ge microstructures, extracted via skeleton graph representation and embedded with a GCN, were analyzed using PCA. In the PCA, the wall structures produced by varying the irradiation angle demonstrated good class separation, achieving a DBI of 1.23. Moreover, changing the irradiation angle resulted in better separation performance than altering the ion fluence, indicating more significant structural transformations.
\section{Acknowledgments}

Some of the Python code in this manuscript was written with the help of the OpenAI ChatGPT o4-mini-high model. The manuscript was also drafted using OpenAI ChatGPT o4-mini-high and DeepL, and the language was reviewed by Enago. After using these tools and services, the authors reviewed and edited the content as needed and take full responsibility for the article.

\small \bibliographystyle{ieeenat_fullname} \bibliography{main}

\begin{thebibliography}{24}
\providecommand{\natexlab}[1]{#1}
\providecommand{\url}[1]{\texttt{#1}}
\expandafter\ifx\csname urlstyle\endcsname\relax
  \providecommand{\doi}[1]{doi: #1}\else
  \providecommand{\doi}{doi: \begingroup \urlstyle{rm}\Url}\fi


\bibitem[Haralick et~al.(1973)Haralick, Shanmugam, and Dinstein]{haralick1973textural}
R.~M. Haralick, K. Shanmugam, and I. Dinstein.
\newblock Textural features for image classification.
\newblock \emph{IEEE Transactions on Systems, Man, and Cybernetics}, SMC-3\penalty0 (6):\penalty0 610--621, 1973.

\bibitem[Ojala et~al.(2002)Ojala, Piet{\"a}ik{\"a}inen, and M{\"a}enp{\"a}{\"a}]{ojala2002multiresolution}
Timo Ojala, Matti Piet{\"a}ik{\"a}inen, and Timo M{\"a}enp{\"a}{\"a}.
\newblock Multiresolution gray-scale and rotation invariant texture classification with local binary patterns.
\newblock \emph{IEEE Transactions on Pattern Analysis and Machine Intelligence}, 24\penalty0 (7):\penalty0 971--987, 2002.

\bibitem[He et~al.(2016)He, Zhang, Ren, and Sun]{he2016deep}
Kaiming He, Xiangyu Zhang, Shaoqing Ren, and Jian Sun.
\newblock Deep residual learning for image recognition.
\newblock In \emph{Proceedings of the IEEE Conference on Computer Vision and Pattern Recognition (CVPR)}, pages 770--778, 2016.

\bibitem[Simonyan and Zisserman(2015)]{Simonyan2015vgg}
Karen Simonyan and Andrew Zisserman.
\newblock Very deep convolutional networks for large-scale image recognition.
\newblock In \emph{Proceedings of the International Conference on Learning Representations (ICLR)}, 2015.
\newblock Oral Presentation.

\bibitem[Edelsbrunner et~al.(2002)Edelsbrunner, Letscher, and Zomorodian]{edelsbrunner2002topological}
Herbert Edelsbrunner, David Letscher, and Afra Zomorodian.
\newblock Topological persistence and simplification.
\newblock \emph{Discrete \& Computational Geometry}, 28\penalty0 (4):\penalty0 511--533, 2002.

\bibitem[Blum(1967)]{blum1967transformation}
Harry Blum.
\newblock A transformation for extracting new descriptors of shape.
\newblock Technical Report Technical Report 216, MIT Research Laboratory of Electronics, 1967.

\bibitem[Kipf and Welling(2017)]{kipf2017semi}
Thomas~N. Kipf and Max Welling.
\newblock Semi-supervised classification with graph convolutional networks.
\newblock In \emph{Proceedings of the 5th International Conference on Learning Representations (ICLR)}, 2017.

\bibitem[Pearson(1901)]{Pearson1901}
Karl Pearson.
\newblock On lines and planes of closest fit to systems of points in space.
\newblock \emph{Philosophical Magazine}, 2:\penalty0 559--572, 1901.

\bibitem[Oishi and Nitta(2018)]{oishi2018nanoporous}
Tomoya Oishi and Noriko Nitta.
\newblock Nanoporous structure formation on the surface of ge by ion beam irradiation.
\newblock \emph{Japanese Journal of Applied Physics}, 57:\penalty0 0913101--1--0913101--6, 2018.

\bibitem[Cao et~al.(2017)Cao, Simon, Wei, and Sheikh]{cao2017openpose}
Zhe Cao, Tomas Simon, Shih-En Wei, and Yaser Sheikh.
\newblock Openpose: Realtime multi-person 2d pose estimation using part affinity fields.
\newblock In \emph{Proceedings of the IEEE Conference on Computer Vision and Pattern Recognition (CVPR)}, pages 7291--7299, 2017.

\bibitem[Yan et~al.(2018)Yan, Xiong, and Lin]{yan2018spatial}
Sijie Yan, Yuanjun Xiong, and Dahua Lin.
\newblock Spatial temporal graph convolutional networks for skeleton-based action recognition.
\newblock In \emph{Proceedings of the Thirty-Second AAAI Conference on Artificial Intelligence (AAAI-18)}, pages 747--755, 2018.

\bibitem[Jivani and Wodo(2022)]{Jivani2022skeletal}
Devyani Jivani and Olga Wodo.
\newblock Skeletal-based microstructure representation and featurization through descriptors.
\newblock \emph{Computational Materials Science}, 214:\penalty0 111668, 2022.

\bibitem[Hu et~al.(2016)Hu, Ziehmer, Wang, and Lilleodden]{hu2016nanoporousgold}
Kaixiong Hu, Markus Ziehmer, Ke Wang, and Erica~T. Lilleodden.
\newblock Nanoporous gold: 3d structural analyses of representative volumes and their implications on scaling relations of mechanical behaviour.
\newblock \emph{Philosophical Magazine A: Journal of Theoretical, Experimental and Applied Physics}, 96\penalty0 (32--34):\penalty0 3322--3335, 2016.

\bibitem[Thomas et~al.(2023)Thomas, Durmaz, Alam, Gumbsch, Sack, and Eberl]{Thomas2023}
Akhil Thomas, Ali~Riza Durmaz, Mehwish Alam, Peter Gumbsch, Harald Sack, and Chris Eberl.
\newblock Materials fatigue prediction using graph neural networks on microstructure representations.
\newblock \emph{Scientific Reports}, 13\penalty0 (1):\penalty0 12562, 2023.

\bibitem[Vecchio et~al.(2021)Vecchio, Mahler, Hammig, and Kotov]{Vecchio2021}
Drew~A. Vecchio, Samuel~H. Mahler, Mark~D. Hammig, and Nicholas~A. Kotov.
\newblock Structural analysis of nanoscale network materials using graph theory.
\newblock \emph{ACS Nano}, 15\penalty0 (8):\penalty0 12847--12859, 2021.

\bibitem[Bradski(2000)]{bradski2000opencv}
Gary Bradski.
\newblock The opencv library.
\newblock \emph{Dr. Dobb's Journal of Software Tools}, 25\penalty0 (11):\penalty0 120--125, 2000.

\bibitem[van~der Walt et~al.(2014)van~der Walt, Sch{\"o}nberger, Nunez-Iglesias, Boulogne, Warner, Yager, Gouillart, Yu, and the scikit-image contributors]{vandewalt2014scikitimage}
St{\'e}fan van~der Walt, Johannes~L. Sch{\"o}nberger, Juan Nunez-Iglesias, Fran{\c c}ois Boulogne, Joshua~D. Warner, Neil Yager, Emmanuelle Gouillart, Tony Yu, and the scikit-image contributors.
\newblock scikit-image: Image processing in python.
\newblock \emph{PeerJ}, 2:\penalty0 e453, 2014.

\bibitem[Zhang and Suen(1984)]{zhang_suen_1984}
T.~Y. Zhang and C.~Y. Suen.
\newblock A fast parallel algorithm for thinning digital patterns.
\newblock \emph{Communications of the ACM}, 27\penalty0 (3):\penalty0 236--239, 1984.

\bibitem[Wang et~al.(2018)Wang, Yan, and Wei]{wang2018imagepy}
A. Wang, X. Yan, and Z. Wei.
\newblock Imagepy: an open-source, python-based and platform-independent software package for bioimage analysis.
\newblock \emph{Bioinformatics}, 34\penalty0 (18):\penalty0 3238--3240, 2018.

\bibitem[YXDragon(2025)]{yxdragon_sknw_2025}
YXDragon.
\newblock {sknw}: skeleton analysis in python.
\newblock \url{https://github.com/Image-Py/sknw}, 2025.

\bibitem[Hunter(2007)]{hunter2007matplotlib}
John~D. Hunter.
\newblock {Matplotlib: A 2D Graphics Environment}.
\newblock \emph{Computing in Science \& Engineering}, 9\penalty0 (3):\penalty0 90--95, 2007.

\bibitem[Fey and Lenssen(2019)]{fey2019fast}
Michael Fey and Jan~E. Lenssen.
\newblock Fast graph representation learning with pytorch geometric.
\newblock In \emph{ICLR Workshop on Representation Learning on Graphs and Manifolds}, 2019.

\bibitem[Pedregosa et~al.(2011)Pedregosa, Varoquaux, Gramfort, Michel, Thirion, Grisel, Blondel, Prettenhofer, Weiss, Dubourg, Vanderplas, Passos, Cournapeau, Brucher, Perrot, and Duchesnay]{pedregosa2011scikit}
Fabian Pedregosa, Ga{\"e}l Varoquaux, Alexandre Gramfort, Vincent Michel, Bertrand Thirion, Olivier Grisel, Mathieu Blondel, Peter Prettenhofer, Ron Weiss, Vincent Dubourg, Jake Vanderplas, Alexandre Passos, David Cournapeau, Matthieu Brucher, Max Perrot, and {\'E}douard Duchesnay.
\newblock Scikit-learn: Machine learning in python.
\newblock \emph{Journal of Machine Learning Research}, 12:\penalty0 2825--2830, 2011.

\bibitem[Davies and Bouldin(1979)]{davies1979cluster}
David~L. Davies and Donald~W. Bouldin.
\newblock A cluster separation measure.
\newblock \emph{IEEE Transactions on Pattern Analysis and Machine Intelligence}, 1\penalty0 (2):\penalty0 224--227, 1979.

\end{thebibliography}

\end{document}